\pdfoutput=1

\documentclass[11pt]{article}

\usepackage[preprint]{acl}

\usepackage{times}
\usepackage{latexsym}

\usepackage[T1]{fontenc}

\usepackage[utf8]{inputenc}

\usepackage{microtype}

\usepackage{inconsolata}

\usepackage{graphicx}

\usepackage{caption}
\usepackage[separate-uncertainty=true]{siunitx}
\usepackage{mathtools}
\usepackage{graphicx}
\usepackage{courier}
\usepackage{epstopdf}
\usepackage{color}
\usepackage{csquotes}

\usepackage{amsmath}
\usepackage{amsfonts}
\usepackage{amssymb}
\usepackage[subnum]{cases}
\usepackage{booktabs} 
\usepackage{pifont}

\usepackage{amssymb}
\usepackage{pifont}

\usepackage {tikz}
\usetikzlibrary {positioning}
\definecolor {processblue}{cmyk}{0.96,0,0,0}
\usepackage{lipsum,adjustbox}

\usepackage[subnum]{cases}
\usepackage{color,soul}

\usepackage{subcaption}

\usepackage{footnote}
\makesavenoteenv{tabular} 

\usepackage{multirow}
\usepackage{booktabs,subcaption,amsfonts,dcolumn}

\usepackage{xcolor}

\usepackage{textcomp}

\usepackage{bbold}
\usepackage{mathtools}
\usepackage{breqn}

\usepackage{amsthm}

\usepackage{makecell}

\usepackage[many]{tcolorbox}
\newtcolorbox{boxA}{
    fontupper = \bf,
    boxrule = 1.5pt,
    colframe = blue 
}

\title{Taming LLMs with Negative Samples: A Reference-Free Framework to Evaluate Presentation Content with Actionable Feedback}

\author{
 \textbf{Ananth Muppidi\textsuperscript{1,2,+}},
 \textbf{Tarak Das\textsuperscript{1,3,+}},
 \textbf{Sambaran Bandyopadhyay\textsuperscript{1}},
 \textbf{Tripti Shukla\textsuperscript{1}},
 \textbf{Dharun D A\textsuperscript{1,3,+}}
\\
\textsuperscript{1}Adobe Research,
 \textsuperscript{2}IIIT Hyderabad,
 \textsuperscript{3}IIT Madras\\
 \small{
   \textbf{Correspondence:} \href{sambaranb@adobe.com}{sambaranb@adobe.com}
 }
}

\begin{document}
\maketitle
\def\thefootnote{+}\footnotetext{Ananth, Tarak and Dharun were interns at Adobe Research when the work was conducted.}
\begin{abstract}
The generation of presentation slides automatically is an important problem in the era of generative AI. This paper focuses on evaluating multimodal content in presentation slides that can effectively summarize a document and convey concepts to a broad audience. We introduce a benchmark dataset, RefSlides, consisting of human-made high-quality presentations that span various topics. Next, we propose a set of metrics to characterize different intrinsic properties of the content of a presentation and present REFLEX, an evaluation approach that generates scores and actionable feedback for these metrics. We achieve this by generating negative presentation samples with different degrees of metric-specific perturbations and use them to fine-tune LLMs. This reference-free evaluation technique does not require ground truth presentations during inference. Our extensive automated and human experiments demonstrate that our evaluation approach outperforms classical heuristic-based and state-of-the-art large language model-based evaluations in generating scores and explanations.
\end{abstract}

\section{Introduction}
Generative AI has opened the scope for several new applications with multimodal data. Recently, the automatic generation of presentation slides given an intent from a user or a long document has gained interest from the research community and industry \citep{sun-etal-2021-d2s,mondal2024presentations,bandyopadhyay-etal-2024-enhancing-presentation,maheshwari2024presentations}. Due to the rapidly increasing volume of work in this area, it is extremely important to systematically assess the quality of the content of such a presentation and provide intuitive, actionable feedback (or explanation), which can justify the scores and helps to improve it further. Manually evaluating presentations needs domain expertise, and is expensive and not scalable. Thus, we aim to come up with an automated system for this task. 

Evaluation has long been a crucial area of research in the processing and generation of natural language. In general, evaluation techniques can be classified as reference-based or reference-free \citep{deutsch2022limitations}. 
Reference-free evaluation is more challenging since there is no reference ground truth to compare the quality of generation during the inference. To address that, machine learning models often rely on contrastive learning frameworks by separating negative samples from positive ones during training \citep{xuan2020hard,wang2022language}.
LLMs as evaluators have recently been used to assess the quality of generation \citep{liu2023gevalnlgevaluationusing,fan2024evascoreevaluationlongformsummarization}. However, there are known facts about several shortcomings of LLMs as evaluators \citep{lan2024criticeval}. For example, the scores obtained from G-Eval \citep{liu2023gevalnlgevaluationusing} can change over different iterations, the range of the output scores is also very limited, and the feedback they can provide is very generic and not actionable. 
Recently, \citet{doddapaneni-etal-2024-finding} showed that current LLM evaluators failed to identify quality drops in generation in more than 50\% of cases on average, with worse performance in the reference-free evaluation. More inconsistencies on direct LLM evaluators are found in other downstream tasks such as story summarization \citep{subbiah-etal-2024-storysumm} and coherence \citep{mu-li-2024-generating}.

Specifically, the task of automatic evaluation with actionable feedback on presentation is challenging due to the following reasons.
(i) There are several aspects on which the content of a presentation can be evaluated. 
(ii) The task is inherently subjective in nature. Deriving rules by looking at a bunch of good presentations is not feasible.
(iii) Training models to evaluate presentation needs dataset with both positive and negative samples. To the best of our knowledge, there is no publicly available dataset that has presentations on a diverse set of academic and non-academic topics. SciDuet \cite{sun-etal-2021-d2s} is a popular dataset that has presentations from research papers on machine learning and NLP.
(iv) It is even more difficult to obtain labeled negative samples for presentations. Moreover, one needs to have a negative samples of different orders to train an evaluation model efficiently.
(v) Finally, it is not possible to have a ground truth good presentation while evaluating a generated presentation during the inference. Thus, a practical evaluation system needs to be reference free in nature for evaluation of presentation with actionable feedback.

Preliminary research has explored the use of simple heuristic-based summarization metrics for document-to-presentation transformation \cite{mondal2024presentations,bandyopadhyay-etal-2024-enhancing-presentation}. However, presentations can be generated from scratch, without reference to any document, and possess distinct narrative properties that separate them from flat summaries. The direct use of LLM as evaluators has several drawbacks, as discussed above.
In this paper, we address the above research gaps by proposing a novel framework to evaluate the content of a given presentation, with the option to consider the original long document, if available. The following are the contributions we have made.
\textbf{(1)} We propose a benchmark dataset comprising $8111$ high-quality presentations from a wide range of topics for training and/or evaluation, referred to as \textit{RefSlides}.
\textbf{(2)} We characterize a set of metrics for the multimodal content of a presentation.
\textbf{(3)} We propose a novel evaluation approach REFLEX (\underline{Ref}erence Free Evaluation with Actiona\underline{l}ble \underline{Ex}planation) that uses perturbations of different degrees to generate negative examples and assesses a presentation content by providing scores with actionable feedback (or explanation).


\section{Problem Statement}\label{sec:prob}
Here, we will present the problem setup formally.
We are given with an unsupervised training set $\mathcal{T} = \{(D_1,P_1), (D_2,P_2),\cdots,(D_n,P_n)\}$ consisting of a set of presentation $\mathcal{P} = \{P_1,P_2,\cdots,P_n\}$ and the corresponding optional source documents $\mathcal{D} = \{D_1,D_2,\cdots,D_n\}$. It can be assumed that a presentation $P_i$ is generated from the source document $D_i$, $\forall i \in [N]$ by summarizing the key topics of the document and with a coherent flow of information by some human experts. When the corresponding source document is not present, we can assume that the presentation is prepared from any source of knowledge (internet for example). We call this data $\mathcal{T}$ unsupervised since we only have positive samples (i.e., human generated presentations) while our task is to evaluate the quality of presentations of different qualities.
Further, a presentation $P_i = (S_{i,1},S_{i,2},\cdots,S_{i,m_i})$ is a sequence (ordered) of $m_i$ slides. A slide contains a slide title and optionally text and images. Number of slides can be different in different presentations. Similarly, a document $D$ can be considered as a sequence of sections and subsections with text and images, as can be seen typically in generally available web documents such as research papers.

During inference, for a presentation $\bar{P}$ and an optional source document $\bar{D}$ coming from a test set $\bar{\mathcal{T}}$, we want to evaluate the quality of the presentation with respect to the set of metrics discussed previously: $\mathcal{M} = \{$Coverage (applicable when the source document is present), Redundancy, Text-Image Alignment, Flow$\}$. For each metric, we will generate a real number score $s \in [0,1]$ and an actionable feedback or explanation $e$ which is a text description about what went wrong in the presentation and the reason for the score. 

\section{Datasets: SciDuet and RefSlides}\label{sec:dataset}

To our knowledge, there is no publicly available data set that has a source document, generated presentations, and the corresponding scores with feedback reflecting the quality of the presentation. So, we rely on a publicly available dataset of document-presentation pairs and curate one ourselves.

We use SciDuet \cite{sun-etal-2021-d2s} which has $1,088$ research publications and their corresponding researcher-made presentations from recent years' NLP and ML conferences (e.g., ACL). We use the same train-test-dev split as mentioned in the SciDuet paper \cite{sun-etal-2021-d2s}. Since these presentations are made by the author of the papers, they can be considered of good quality.

Since SciDuet only has scientific presentations, we propose a new dataset, referred to as \textbf{RefSlides}, consisting of $8,111$ presentations in diverse domains such as marketing, business, education, healthcare, etc. all collected from SlideShare\footnote{\url{https://www.slideshare.net}}. We carried out a train-eval-test split of $70:15:15$ on RefSlides, resulting in $1,216$ presentations in each of the test and eval splits.
We initially downloaded 17,595 unique presentations. 
However, not all SlideShare slides are of really good quality. To be able to select only positive samples for our training pipeline, we have performed a detailed preprocessing and used multiple filters on the available presentations, summarized as follows. (1) The number of slides should be between 5 and 35. (2) An introductory and a concluding slide should be there. (3) Two consecutive slides should not have more than $80\%$ overlap. (4) The language should be English. And (5) The aspect ratio of the presentation should strictly be 16:9. Finally, we also manually scrutinize the presentations to ensure their quality. The above filtering retains only $8,111$ presentations ($46\%$) which are of good quality. More details about the dataset can be found in Appendix \ref{appen:RefSlides}.


\section{Proposed Metrics}\label{sec:metrics}
We focus on the following four metrics to evaluate a presentation. The implementation of these metrics through a novel contrastive learning framework is discussed in the next section. The list of metrics is not exhaustive, and one can include new metrics in our framework with some additional work.

\subsection{Coverage}
This metric is applicable when the source document is given along with the generated presentation~\footnote{Please note that source document is available in specific usecases such as making a presentation from a research paper. However, for broader applications such as sales and marketing, presentations are made from scratch}. We define coverage as a measure of how well crucial information from the source document is captured in the slides. We include this metric because the main objective of document-to-presentation transformation is effortless knowledge transfer, and the presentation should cover most of the important aspects of the document. Please note that existing works use simple cosine similarity based heuristics \citep{maheshwari2024presentations,bandyopadhyay-etal-2024-enhancing-presentation} or LLM driven question-answering mechanisms to measure coverage \citep{deutsch2021towards}. Typically, these heuristics do not capture the notion of `importance' as they are averaged over all possible content pairs from the document and the presentation. LLM-based question-answer generation mechanisms often lead to hallucination and lack interpretability.

\subsection{Redundancy}
We define redundancy as the existence of redundant material in the slides. This metric was included as it is important to keep the presentation concise and not provide the same information repeatedly. There are simple heuristics to measure redundancy using the average cosine similarity of content within a document or presentation \citep{maheshwari2024presentations}. However, because of the average on the entire content, the metric becomes diluted. 

\subsection{Text-Image Alignment}
We define this metric as a measure of how well an image corresponds to the text in the slide. This metric is important for several reasons. First, it improves understanding by visually reinforcing key points with relevant images, making the content more engaging and memorable. Second, it helps clarify complex concepts or data by providing visual explanations that complement textual information. The cosine similarity of multimodal embeddings, such as CLIP embeddings \citep{radford2021learning} of text and images, can be used as a heuristic to compute this metric. However, images within a presentation (or a document) on a specific topic can be very similar to each other, and embedding-based cosine similarity is not enough to understand the relevance of text and images in this context. In our approach, we sample negative images from the same presentation, making it suitable to compute this metric effectively.

\subsection{Flow}
We define flow as a logical progression of content from one point to another and organizing information in a way that maintains a smooth transition between ideas, topics, or sections within the slides, thereby enhancing understanding and retention. The flow of content in a presentation is crucial because it guides the audience through the information in a logical and coherent manner, thus governing the overall narrative \citep{yang2023doc}. Please note that there are works on evaluating the narrative of a story \citep{tian-etal-2024-large-language} and movies \citep{su-etal-2024-unveiling}, but our work is the first to evaluate the flow (or narrative) of a presentation.

\section{Detailed Pipeline of REFLEX}\label{sec:REFLEX}

\subsection{Content Extraction and Summarization}\label{sec:preprocessing}
First, we explain the extraction of multimodal content from a presentation and the source document if present, and the preprocessing of them.
\subsubsection{For Presentations}
Due to the challenges associated with processing slide images, particularly the need for OCR or LMMs at each step, we first perform feature extraction on each slide image before passing the data to downstream sub-modules. The extracted features include the slide's text content and text descriptions of any images present. For this purpose, we utilize the Phi-3-vision-128k-instruct model\cite{abdin2024phi3technicalreporthighly}, the current state-of-the-art LMM publicly available. Each set of extracted features is then summarized into descriptions for the slide title, text summary, image caption, and image description in a few lines. The final pre-processed object for a presentation from this step is an array of JSON elements each of the format.
\begin{boxA}
\footnotesize
$S_{i,j}$ = \{`title': \textit{<Title of Slide>},\\ `text': \textit{<Short summary of text>},\\ `image-caption': \textit{<Caption(s) of the image(s)>},\\ `image': \textit{<Text description of the image(s)>}\}.
\end{boxA}
Hence, each presentation object will be:\\
$P_i = [S_{i,1}, S_{i,2}, \cdots, S_{i,m_i}] \forall i \in [N]$ where $S_{i,j}$, $\forall j$ are in the same reading order as in the original presentation.


\begin{figure}[h]
    \centering
    \includegraphics[width=\linewidth]
    {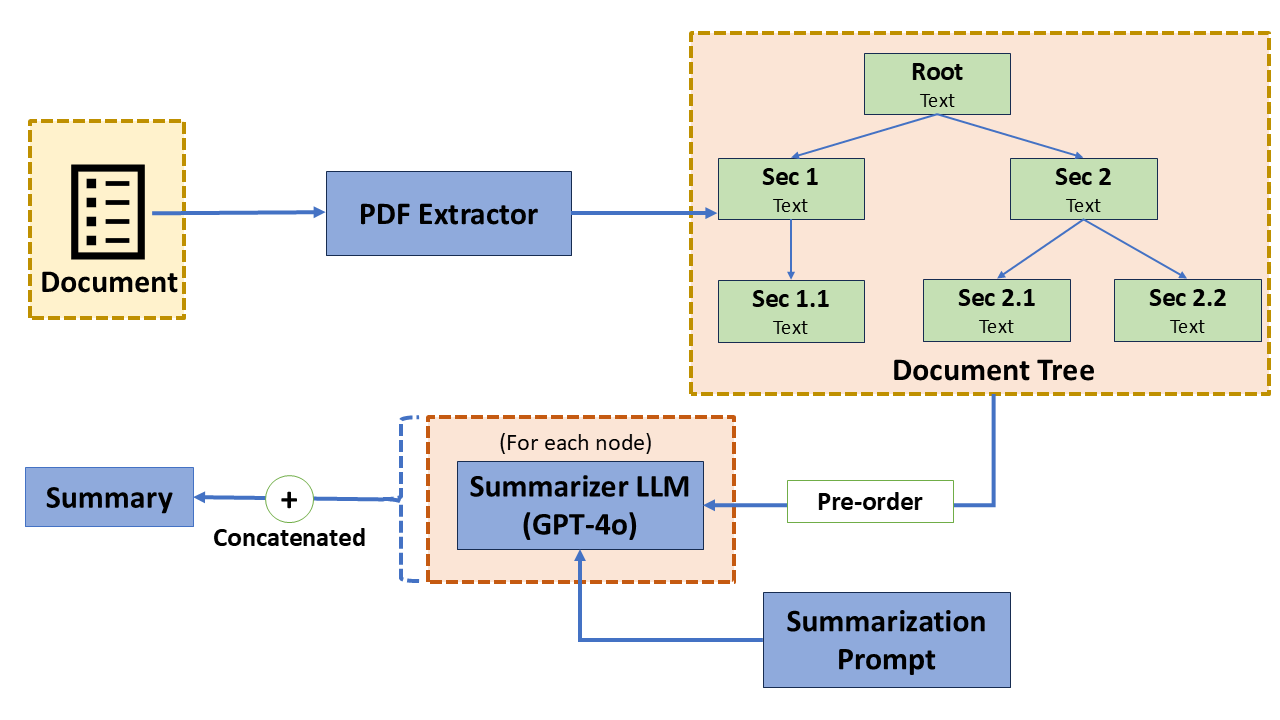}
    \caption{Overview of the architecture of document preprocessing and summarization sub-module of REFLEX.}
    \label{fig:summary_arc}
\end{figure}

\subsubsection{For Documents}
Additionally, if a document is present, it is hierarchically summarized using an LLM to retain as much information as possible. First, we use a publicly available PDF Extract API\footnote{\url{https://developer.adobe.com/document-services/apis/pdf-extract}} to extract the section-wise text content from the documents. The hierarchical summarization is then carried out by re-organising the extracted document into a tree as shown in Fig. \ref{fig:summary_arc} and then doing a pre-order traversal to summarise each of the sections using GPT-4o (prompt is given in Appendix \ref{appen:sum_prompt}) followed by concatenation of the sub-section summaries. Summarization based on pre-order traversal preserves the important information even from the leaves of the document tree, preventing losses \citep{bandyopadhyay-etal-2024-enhancing-presentation}. This step is crucial because LLMs have context-length limitations, and we observe a significant performance decline in terms of information content as context length increases.

\begin{figure}[h]
    \centering
    \includegraphics[width=\linewidth]
    {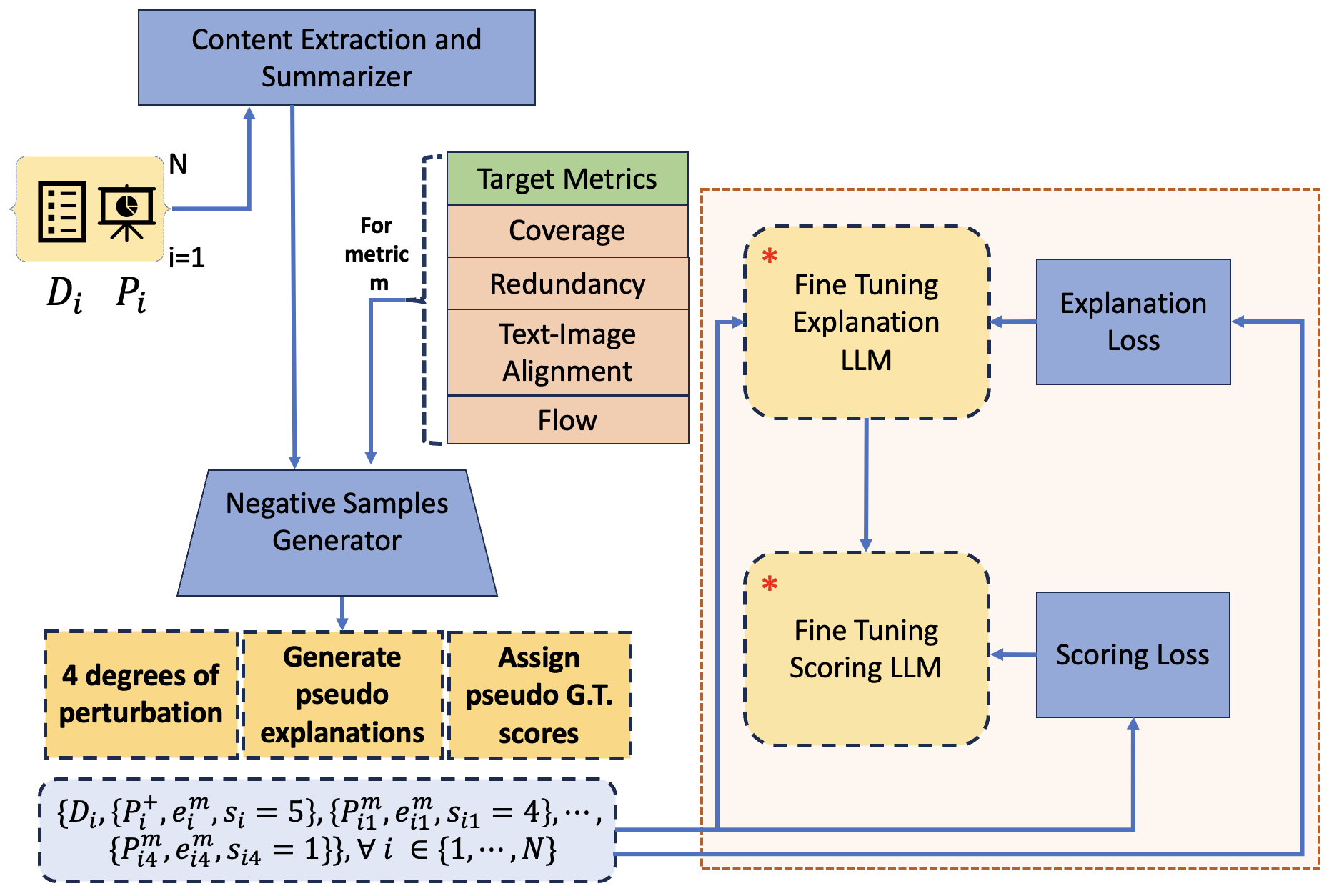}
    \caption{Training pipeline of REFLEX (best seen in colors)}
    \label{fig:reflex_training}
\end{figure}

\subsection{Generating Negative Samples of Different Degrees}\label{sec:genNegSamples}
As discussed before, to effectively train a model that can separate different qualities of presentations from each others and score them accordingly, we need presentations of different qualities with ground truth scores reflecting the qualities. It is possible to obtain human-made presentations, which after a certain checks and pre-processing (as discussed in Section \ref{sec:dataset}), can be considered of very good qualities (positive samples). However, presentations with varying amounts of flaws and the corresponding ground truth scores are not possible to obtain from the available sources. Moreover, we also need a precise explanation of why some presentation is not good for generating the actionable feedback along with the scores.
To achieve both, we apply controllable and carefully crafted perturbation techniques of different degrees to a good quality presentation (positive sample) to generate negative presentations along with the corresponding scores and explanations to complete the training data. In this work, we consider perturbations up to degree 4 and the ground truth scores is in the range of 1 to 5. The higher the degree of perturbation, there are more changes in the presentation.

More formally, we consider the presentations from the given training set $\mathcal{T} = \{(D_1,P_1), (D_2,P_2),\cdots,(D_n,P_n)\}$ as positive samples and denote them by $P_i^+$ instead of $P_i$, $\forall i$. We propose a perturbation function $h_d^m: (D_i, P_i^+) \mapsto (P_{id}^m, e_{id}^m, s_{id})$, corresponding to each metric $m \in \mathcal{M}$ and for each degree $d$, and map a positive presentation $P_i^+$ (with the optional source document $D_i$) to a negative presentation $P_{id}^m$ of degree $d$. In addition, it also generates the corresponding pseudo-ground truth explanation (actionable feedback) $e_{id}^m$ and the pseudo-ground truth score $s_{id}$. Please note that $d \in \{1,2,3,4\}$ since we consider only four degrees of perturbations. We also set the pseudo ground truth score $s_{id} = 5 - d$ for degree perturbation $d$. Thus, for each positive sample, there are four negative samples of different degrees created. For completeness, we add a score of $s_i = 5$, $\forall i \in [N]$ to the positive (original) presentations and a metric-specific explanation $e_i^m$, which will be discussed next.

\subsubsection{Metric Specific Perturbations}
Here, we will discuss the details of the perturbation function $h_d^m$ for each metric $m$ with different degrees $d = \{1,2,3,4\}$.

\textbf{Coverage}:
Perturbations for the metric $m=$ Coverage are created by removing content specific to some randomly selected topics from the presentation.
First, we pass the summarized content from a document $D_i$ to an LLM to GPT4o to obtain a set of topics (let us denote it as $T_i$) associated with the content of the document. Then, we pass the content of each slide to the same LLM and ask it to assign one or more topics from $T_i$.
Based on the degree $d$ of perturbation $h_d^m$, $x\%$ of topics (we set $x = d \times 20$) are selected at random from $T_i$. Then we consider all the slides which are assigned to any of these selected topic and removed those slides from the presentation to obtain $P_{id}^m$. We generate the corresponding pseudo ground truth explanation as $e_{id}^m =$ \textit{``The following topics from the source document should be added: <Topic 1>, <Topic 2> . . .''}. For a positive slide $P_i^m$, the corresponding explanation $e_i^m$ is \textit{``All major topics from the source document are covered in this presentation''}.

\textbf{Redundancy}:
Perturbations for the metric $m=$ Redundancy are created by adding duplicate content within the presentation. Based on the degree $d$ of perturbation $h_d^m$, $x\%$ of slides (again, we set $x = d \times 20$) are selected at random, duplicated, and inserted at random positions within the presentation. We generate the corresponding pseudo ground truth explanation as $e_{id}^m =$ \textit{``The following topics in the slide seem to be repeated: <slide_idx_1>. <Topic 1>, <slide_idx_2>. <Topic 2> . . .''}. For a positive slide $P_i^m$, the corresponding explanation $e_i^m$ is \textit{``Slides are concise and there is little to no redundant information''.}

\textbf{Text-Image Alignment}:
Perturbations for the metric $m=$ Text-Image Alignment are created by interchanging the images within the slides of the presentation.
Based on the degree $d$ of perturbation $h_d^m$, $x\%$ of slides (where, $x = d \times 20$) with images are selected at random. Then, we create a random permutation of the positions of the selected slides. The images from each of the selected slides are removed and placed to the corresponding slides in the permutation to obtained the perturbed sample.
For example, let the positions of selected slides within the presentation be [1,3,7] and assume the random permuted order of this list is [3,7,1]. Then the images in the 1st slide of $P_{i}^+$ will now be placed in the 3rd slide of $P_{id}^m$. Similarly, images of the 3rd slide of $P_{i}^+$ will be in 7th slide of $P_{id}^m$ and so on.
We generate the corresponding pseudo ground truth explanation as $e_{id}^m =$ \textit{"The following slides seem to have images misaligned with the text: <slide_idx>. <Topic 1>, <slide_idx>. <Topic 2> . . ."}. For a positive slide $P_i^m$, the corresponding explanation $e_i^m$ is \textit{``All the slides have images relevant to their text''}.

\textbf{Flow}:
Perturbations for the metric $m=$ Flow are created by changing the flow of content within the presentation. 
Based on the degree $d$ of perturbation $h_d^m$, $x\%$ of slides (where, $x = d \times 20$) are selected at random and the positions of the selected slides are permuted randomly.
For example, let the positions of selected slides within the presentation be $[1,3,7]$ and assume the random permuted order of this list is $[3,7,1]$. So, $S_{id,3}^m = S_{i,1}^+$, $S_{id,7}^m = S_{i,3}^+$, and $S_{id,1}^m = S_{i,3}^+$, where $S_{id,3}^m$ is the 3rd slide of the perturbed presentation $P_{id}^m$ and $S_{i,1}^+$ is the first slide of the original (positive) presentation $P_{i}^+$, and so on.
We generate the corresponding pseudo ground truth explanation as $e_{id}^m =$ \textit{"The progression of topics would be more coherent if the slide were ordered: <slide_idx>. <Topic 1>, <slide_idx>. <Topic 2> . . ."}. For a positive slide $P_i^m$, the corresponding explanation $e_i^m$ is \textit{``The presentation follows a natural, coherent flow''}.

At the end of this, for each metric $m$, we have an updated training dataset $\hat{\mathcal{T}}_m = \bigl\{D_i, (P_i^+, e_i^m, s_i), \{(P_{id}^m,e_{id}^m, s_{id})\}_{d=1}^4 \bigr\}_{i=1}^N$, where the first entry is the optional source document, second entry is the positive sample and other four examples are negative samples with different degrees of perturbations. All the positive and negative samples have their corresponding explanations and scores. Thus, the training set for each metric is $5$ times than the original training set provided in Section \ref{sec:prob}.


\subsubsection{Training LLM to Generate Explanation}\label{sec:PGTExpl}
For each of the metrics above, we train a model to generate actionable explanations on how to improve the presentations' quality for that metric. 
We use Phi3-Mini \cite{abdin2024phi3technicalreporthighly} as the base language model for generating explanations.  
To fine-tune this base model, we consider the positive and perturbed presentations (and the source document for the metric coverage) as the inputs and corresponding explanations as the target output from the updated training set $\hat{\mathcal{T}}_m$.
We use standard CrossEntropy loss over the tokens generated by the model and the tokens in the pseudo ground truth explanations.
We use LoRA \cite{hu2021loralowrankadaptationlarge} to fine-tune our base LLM. This choice was motivated by the fact that since there are multiple metrics, we will create multiple explanation modules. Instead of fine-tuning 4 separate language models, we only fine-tune the LoRA adapters so that the adapters can be interchangeably used with the same base model. 
Since the fine-tuning dataset contains explanations about why the perturbed slides are negative, the model implicitly learns what constitutes a good slide. During inference, this model can now be used for generating explanations on how to improve slide quality.

\subsubsection{Training LLM to Generate Score}\label{sec:PGTScore}
Here also we aim to use Phi3-Mini \cite{abdin2024phi3technicalreporthighly} with a single neuron regression head to output a score between 0 and 1, and utilize the same LoRA fine-tuning approach as followed for generating explanations.
The input to this model is the presentations (both positive and negative samples from $\hat{\mathcal{T}}_m$ and use the corresponding fine-tuned model generated explanations from Section \ref{sec:PGTExpl}. The target output are the corresponding scores from $\hat{\mathcal{T}}_m$. We use the MSE loss to train the network. 

Thus, we obtain two fine-tuned models (same base model with different LoRA adapters) for each metric. The whole training process is depicted in Figure \ref{fig:reflex_training}.



\subsection{Inference}
During inference for a selected metric, the input is a presentation, and the source document if the selected metric is coverage. We first preprocess the presentation and the source document as discussed in Section \ref{sec:preprocessing}. Next, we feed the processed presentation (and the document if applicable) to the fine-tuned explanation generation model which generates the explanation. Then, the same input along with the generated explanation is fed to the fine-tuned scoring model to generate the score. The inference process is depicted in Figure \ref{fig:reflex_inference}.

\begin{figure*}[h]
    \centering
    \includegraphics[width=\textwidth]{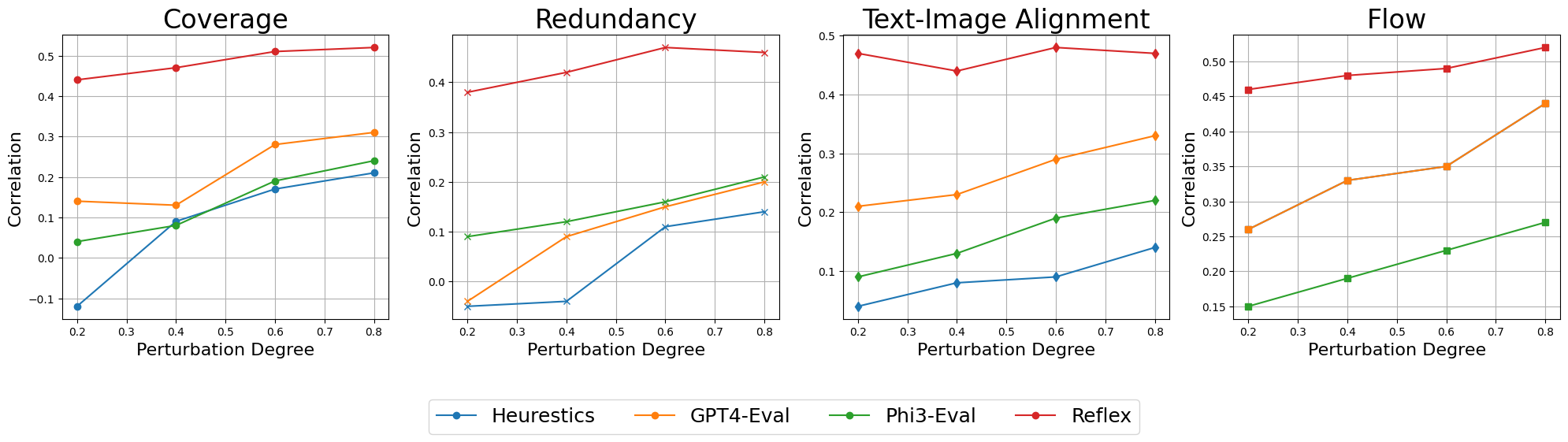}
    \caption{Correlations between the pseudo ground truth and metric for each degree. Reflex shows a significantly higher correlation overall, while also maintaining sensitivity to detecting smaller degrees of perturbations, where other baselines show a performance decline.}
    \label{fig:sample_image}
\end{figure*}

\begin{table}[h]
\centering
\resizebox{1.0\linewidth}{!}{
\renewcommand{\arraystretch}{0.9} 
\begin{tabular}{lcccccc}
\toprule
\textbf{Metric} & \textbf{Method} & \multicolumn{2}{c}{\textbf{SciDuet}} & \multicolumn{2}{c}{\textbf{RefSlides}} \\ 
\cmidrule(lr){3-4} \cmidrule(lr){5-6}
 & & \textbf{$\rho\uparrow$} & \textbf{$\tau\uparrow$} & \textbf{$\rho\uparrow$} & \textbf{$\tau\uparrow$} \\ 
\midrule
\multirow{4}{*}{\textbf{Coverage}}  
  & Heuristic & 0.16 & 0.07 & -- & -- \\ 
  & Phi3-Eval & 0.19 & 0.21 & -- & -- \\ 
  & G-Eval & 0.31 & 0.30 & -- & -- \\ 
  & Reflex & \textbf{0.47} & \textbf{0.49} & -- & -- \\ 
\midrule
\multirow{4}{*}{\textbf{Redundancy}} 
  & Heuristic & 0.18 & 0.19 & 0.16 & 0.19 \\ 
  & Phi3-Eval & 0.13 & 0.20 & 0.07 & 0.11 \\ 
  & G-Eval & 0.38 & 0.33 & 0.31 & 0.28 \\ 
  & Reflex & \textbf{0.51} & \textbf{0.47} & \textbf{0.52} & \textbf{0.49} \\ 
\midrule
\multirow{4}{*}{\textbf{Text-Image Ali.}} 
  & Heuristic & 0.21 & 0.17 & 0.14 & 0.11 \\ 
  & Phi3-Eval & 0.17 & 0.14 & 0.20 & 0.18 \\ 
  & G-Eval & 0.40 & 0.36 & 0.37 & 0.33 \\ 
  & Reflex & \textbf{0.59} & \textbf{0.52} & \textbf{0.49} & \textbf{0.44} \\ 
\midrule
\multirow{3}{*}{\textbf{Flow}}  
  & Phi3-Eval & 0.04 & -0.10 & -0.03 & -0.04 \\ 
  & G-Eval & 0.23 & 0.21 & 0.16 & 0.14 \\ 
  & Reflex & \textbf{0.41} & \textbf{0.47} & \textbf{0.44} & \textbf{0.42} \\ 
\bottomrule
\end{tabular}
}
\caption{Comparison of correlation coefficients between baselines and pseudo ground truths}
\label{tab:table_1}
\end{table}

\section{Experimental Evaluations}\label{sec:exp}

\subsection{Baselines}\label{sec:baselines}
\subsubsection{Heuristic based Baselines}
Heuristic based methods are simple to calculate and have been used in literature \citep{he2023hauserholisticautomaticevaluation,bandyopadhyay-etal-2024-enhancing-presentation}. 

We use the following formula to calculate the heuristic for coverage: $Cov(P,D) = \frac{1}{|D| \cdot |P|} \sum_{x \in D} \sum_{y \in P} cosine(x, y)$. It is essentially the average cosine similarity between the sentence embeddings \citep{reimers-2019-sentence-bert} of the paragraphs within the document and that of slides within the presentation. 
%
Similarly for redundancy, we calculate the average sentence embeddings of any two slides within a presentation and subtract that from 1 as follows: $Redun(P) = 1 - \frac{1}{|P|^2} \sum_{x,y \in P}cosine(x, y)$. If the slides are less similar to each other, more will be the value of this metric.
For text-image alignment, we compute the average sentence embeddings of the text and image description (please refer to Section \ref{sec:preprocessing}) of the text and images present in the same slide. This is calculated as $\frac{1}{|P|} \sum_{S \in P} cosine(Img_P, Text_P)$, here $|P|$ is the number slides containing both text and images. 

We could not find any intuitive heuristic for flow.

\subsubsection{G-Eval}
We have used G-Eval framework \cite{liu2023gevalnlgevaluationusing} with GPT-4o as one of our baselines. In this framework, GPT-4o is tasked to come up with evaluation steps for each metric given a brief description of the metric. These evaluation steps are fed to GPT-4o via another call to get the scores over 128 iterations and we have used the average of the score. To keep the comparaision fair, in this framework, for explanation generation, we use in-context examples to exemplify the expected explanation when prompted with a slide. An example of the exact prompt provided is shown in the appendix.

\subsubsection{Phi3-Eval}
To create a fair comparison, we also use Phi3-Eval, which uses the same framework as G-Eval, but the base language model used is Phi3-Mini, which is the same model used in our methodology. The same in context prompting is also used in this baseline.

\subsection{Experimental Setup}
For every scoring and explanation module, the Phi3-Mini base model, along with LoRA adapters were used. All computation was done on a cluster of two A100-80GB servers. For explanation tuning, cross entropy loss was used, and for score generation, MSE was used for regression (output is a continuous range between [0, 1]). For each training, hyper parameter tuning was conducted via grid search, and the search space is mentioned in the appendix below. For all text generations, \textit{temperature} was set to 0, and sampling was disabled to ensure model determinism.

\subsection{Automatic Evaluations}
\subsubsection{Correlation of Models with Pseudo Scores}
We evaluate different metrics via the Spearman and Kendal-Tau correlations. Table \ref{tab:table_1} shows the correlations of the baselines and Reflex with the assigned Pseudo Ground Truth. The correlations of all metrics indicate that our model correlates the highest with the Pseudo Ground Truth compared to all baselines.

\subsubsection{Performance across perturbation degree}
We compare the Kendall-Tau correlations of the baselines and our model at various degrees of perturbations to visualize the sensitivity of the evaluation model. Ideally, the model should be able to detect various degrees of perturbations, as compared to only being able to detect larger degrees. 

As visualized in Figure \ref{fig:reflex_inference}, our model performs well even at smaller degrees of perturbation, while the baselines display significantly poorer correlation, indicating that our model is sensitive to nuances in the input presentation that the baselines fail to capture.

\subsubsection{Evaluation Explanations}

The ROUGE metric is commonly used for automatic evaluation of summaries \cite{lin2004rouge}.
We compute the ROUGE-F1 scores between the ground truth explanations and the model-generated explanations to evaluate their quality and compare the same to explanations generated by few-shot prompting (to generate explanations with the same syntactic structure) both GPT-4o and Phi3-Mini as baselines. From Table \ref{table:rouge}, our model generated significantly better explanations than the baselines. This can be attributed to the contrastive training methodology which enables our framework to have a more robust understanding of the constitution of a good presentation.

\begin{table} 
\centering
\resizebox{1.0\linewidth}{!}{
\begin{tabular}{lccc}
\toprule
\textbf{ROUGE Metric} & \textbf{Few-Shot GPT-4o} & \textbf{Few-Shot Phi3-Mini} & \textbf{REFLEX} \\
\midrule
\textbf{ROUGE-1$\uparrow$} & 0.472 & 0.341 & \textbf{0.658} \\
\textbf{ROUGE-2$\uparrow$} & 0.284 & 0.132 & \textbf{0.361} \\
\textbf{ROUGE-3$\uparrow$} & 0.131 & 0.087 & \textbf{0.214} \\
\textbf{ROUGE-L$\uparrow$} & 0.411 & 0.314 & \textbf{0.623} \\
\bottomrule
\end{tabular}
}
\caption{Comparison of ROUGE F1 scores between Few-Shot GPT-4o, Few shot Phi3-Mini and Reflex.}
\label{table:rouge}
\end{table}

\subsection{Human Evaluation Results}
We randomly sampled 50 data samples from SciDuet and instructed human annotators (experienced in the field of Machine Learning) to rate the quality of the presentations along the 4 metrics on the 5-point Likert scale ([1,2,3,4,5]). 3 annotators rated each sample. Details about the human annotation is given in the appendix.
\subsubsection{Inter-Annotator Agreement}
Following \cite{he2023hauserholisticautomaticevaluation} we calculate the mean and max of inter-annotator agreement via both Kendalls-Tau and Spearman Correlation by holding out the ratings of one annotator at a time and computing the correlation between the other two. This way, we compute the mean among the three different pairs and also report the maximum correlation between any pair. The correlations are reported in Table \ref{table:inter-anno} in Appendix \ref{appen:inter-anno}.


\subsubsection{Comparison with human rankings}
In Table \ref{table:corr}, we compute the Spearman Correlations and the Kendalls-Tau to evaluate our metric. Our model demonstrates higher correlation than the baselines for all proposed metrics apart from Flow, where it is the second best model. It is worth noting that the interannotator agreement for flow is significantly lower than for all other metrics, suggesting that it is subjective in nature.

\begin{table}[h] 
\centering
\resizebox{0.8\linewidth}{!}{
\small
\begin{tabular}{lcccc}
\toprule
\textbf{Metric} & \textbf{Method} & \multicolumn{2}{c}{\textbf{Corr.}} \\ 
\cmidrule(lr){3-4} 
& & \textbf{$\rho\uparrow$} & \textbf{$\tau\uparrow$} \\

\midrule
\multirow{4}{*}{\textbf{Coverage}}  
  & Heuristic & -0.11 & -0.09 \\ 
  & Phi3-Eval & 0.12 & 0.09 \\ 
  & G-Eval & 0.37 & 0.33 \\ 
  & Reflex & \textbf{0.46} & \textbf{0.41} \\ 
\midrule
\multirow{4}{*}{\textbf{Redundancy}} 
  & Heuristic & -0.04 & 0.03 \\ 
  & Phi3-Eval & 0.16 & 0.11 \\ 
  & G-Eval & 0.42 & 0.38 \\ 
  & Reflex & \textbf{0.52} & \textbf{0.47} \\ 
\midrule
\multirow{4}{*}{\textbf{Text-Image Ali.}} 
  & Heuristic & -0.06 & 0.07 \\ 
  & Phi3-Eval & 0.21 & 0.19 \\ 
  & G-Eval & 0.36 & 0.35 \\ 
  & Reflex & \textbf{0.45} & \textbf{0.43} \\ 
\midrule
\multirow{3}{*}{\textbf{Flow}}  
  & Phi3-Eval & 0.21 & 0.14 \\ 
  & G-Eval & \textbf{0.36} & \textbf{0.33} \\ 
  & Reflex & 0.32 & 0.29 \\ 
\bottomrule
\end{tabular}
}
\caption{Comparison of correlation coefficients between baselines and human annotations}
\label{table:corr}
\end{table}

\begin{figure}[h]
    \centering
    \includegraphics[width=0.6\linewidth]{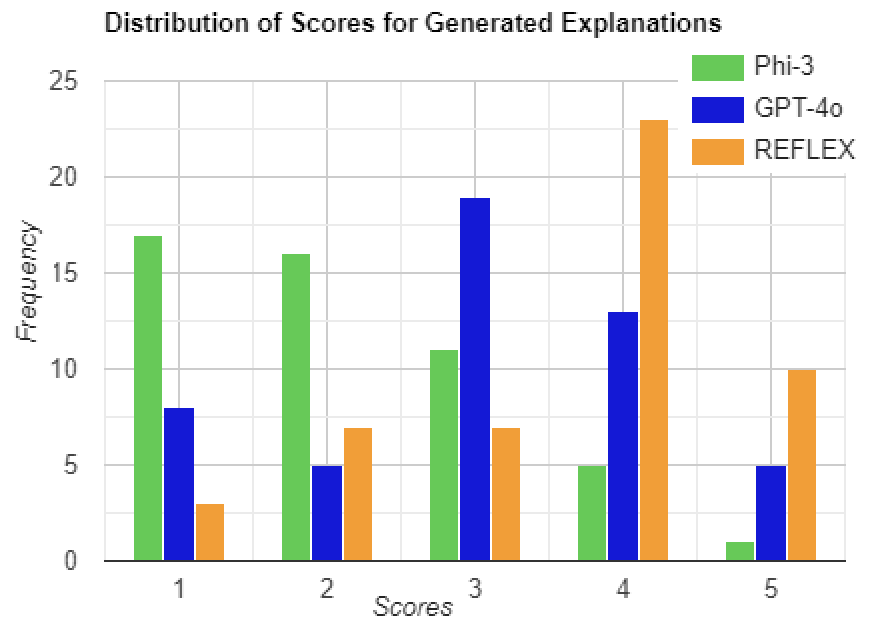}
    \caption{Distribution of scores for explanations generated by various models assigned by human annotators. Higher score is better. As noticeable, REFLEX has highest assignment of 4 and 5.}
    \label{fig:bar-graph}
\end{figure}

\subsubsection{Human evaluation of explanations}
Along with the scores, we also presented the human annotators with explanations generated from our model, along with the baselines G-Eval and Phi3-Eval. Annotators were asked to rate, on a scale of 1 to 5, whether the explanations provided  would improve the quality of the input presentation. For a given presentation. The mean inter-annotator agreement was 0.71 and maximum was 0.78. The distribution of the scores are shown in Figure \ref{fig:sample_image}. The annotation shows that REFLEX, on average scored higher than the other baselines, achieving the highest number of "5" and "4" scores.

\section{Discussions}
In this study, we introduced REFLEX, a method to evaluate the multimodal content quality of generated presentations using four metrics. We proposed a systematic approach for creating perturbed presentations and utilized a Phi-3 fine-tuned LLM base architecture. 
Both automated and human evaluations show that REFLEX outperforms state-of-the-art LLMs, including the larger GPT-4-based G-Eval, despite having fewer parameters.

\section*{Limitations}

Despite the promising results of our framework, there are several limitations to consider:

\begin{itemize}
    \item \textbf{Model size limitations:} We were unable to use larger models such as LLaMA-70B, Mixtral-22B, or other state-of-the-art large language models due to computational and resource constraints. The models used in our experiments are smaller, which may impact the quality of the generated explanations and scores.
    In our work, we show that our framework with a fixed model size (Phi3-Mini) outperforms baselines using a similar model size, but with the availability of compute larger models can be used to bolster performance.
    
    \item \textbf{Non-exhaustive metrics:} The set of metrics used in this framework—\textit{Coverage}, \textit{Redundancy}, \textit{Text-Image Alignment}, and \textit{Flow}—is not exhaustive. Additional metrics, such as engagement, clarity, or overall aesthetic, could provide a more comprehensive evaluation of the presentation slides. Our framework can easily be adapted to include metrics, given that negative samples of varying degrees of perturbation can be carefully designed.
    
    \item \textbf{Lack of design evaluation:} Our framework solely focuses on the content of the slides, without accounting for the visual design aspects such as layout, typography, or color schemes. Future work could incorporate design evaluation metrics in a constrastive learning methodology.
\end{itemize}


\bibliography{REFLEX}

\newpage
\appendix

\section*{Appendix}

\maketitle

\section{More details on RefSlides}\label{appen:RefSlides}
We initially downloaded $17,595$ unique presentations from Slideshare and applied a 3-step filtering process to curate a cross-domain high-quality dataset:
\begin{enumerate}
    \item Length: We worked with presentations of length 5 to 35 slides only. This is because presentations less than 5 slides long were found to be lacking any considerable information and those longer than 35 slides were rejected due to context-length limitations of LMMs.
    \item Heuristics: Based on examination of randomly sampled downloaded presentations, we came up with a set of qualitative criteria to reject low-quality samples: 
    \begin{itemize}
        \item Introductory and Concluding Slides: We assumed a presentation can be called `complete' only if it has an introductory and a concluding slide.
        \item Duplicate Slides: We removed all the presentations from our dataset which had more than 80\% overlap among consecutive slides indicating animations as they might interfere with our metrics.
        \item Language: We kept only slides in English for our experiments.
    \end{itemize}
    \item Aspect Ratio: We strictly limit all the presentations to be of $16:9$ aspect ratio.
\end{enumerate}
For step-2, we convert each presentation $P_i$ into a set of images of slides $P_i = [S_{i,1}, S_{i,2}, \cdots, S_{i,m_i}] \forall i \in [N], P_i \in \mathcal{P}$. This array is then parsed into batched of $5$ slide-images concatenated vertically and fed into GPT-4o using a suitable prompt to evaluate for each of the above criteria in step-2.\\ 
After conducting the steps above, we end up having 8111 high-quality presentations in the dataset.

\section{Prompts}

\subsection{Prompts for Data Extraction}
For the data extraction, Phi3-Vision was prompted for both text and image elements.

\begin{boxA}
\textbf{TEXT_PROMPT} = "I have input an image of a slide from a presentation. Your task is to carefully go through the text content of the slide, and describe the text present in two sentences and two sentences only. Do NOT consider the images on the slide. If there is no text, just return "No Text". Make it concise" 
\end{boxA}

\begin{boxA}
\textbf{IMAGE_PROMPT} = "I have input an image of a slide from a presentation. Your task is to carefully go through the image(s) of the slide, and describe the image(s) present in two sentences and two sentences only (per image). Do NOT consider the text on the slide. If there are no images, just return No Images. If there are multiple images, return Image1: <description>, Image2: <description> and so on. Exclude logos, templates and backgrounds." 
\end{boxA}

\subsection{Prompt for Document Summary}\label{appen:sum_prompt}
For generating hierarchical summary of document, each node of the input document tree along with the following prompt is provided to GPT-4o.
\begin{boxA}
\textbf{SUMMARY_PROMPT= "You are now an expert at generating highly informative summary from a given titled document.\\
Read the document very carefully. The summary should be as informative and coherent as possible.The summary should be balanced. Include the key points and main ideas without adding non-essential information.\\
The goal is to convey the core message and important details in a clear and concise manner.\\
Keep summary for each section in exactly 30 words. For the passage titled "Reference", keep the summary in 20 words.\\
Do NOT mention section title in the summary.\\
<Text in the node>
"}
\end{boxA}

\subsection{Prompts for Topic Extraction}

For coverage, the input document along with the following prompt is provided to GPT-4o to extract topics.
\begin{boxA}
\textbf{TOPIC_PROMPTS} = "I have the summary of a document, as follows: 
\\
{<Input summary>}
\\
From this summary, please extract all major themes / topics in the form of keywords. Be as detailed as possible and extract all relavant topics.
\\

Your output must be in the form of a python list, with each topic separated by commas. Make sure all topics are lowercase, and all spaces are separated by underscores.
" 
\end{boxA}

The model reponse is then parsed as a python array for further use.

\subsection{Prompts for G-Eval and Phi3-Eval}
G-Eval and Phi3-Eval are both used as baselines in our bench marking. The following prompts were used to generate the explanations in our experiments. The prompt includes an in context example so as to match the generation that our models were fine tuned to generate. 

\begin{boxA}
\textbf{PROMPT_G_EVAL} = "
\\
You are a helpful language model and your task is to judge a presentation based on whether it contains redundant information. This can include any repetitive text/images. The input will be in the form of a list of slides, where each slide will contain a textual description of both the text and the image present in the slide.

<input_slide>

Your output must be suggestions on which slides need to be removed. Your output format should strictly follow the template outlined below. For example, if you feel slides with topic 1 and topic 2 need to be removed, include these slides in your response as so: 
\\
\textit{
"The following topics in the slide seem to be repeated: 
\\
1. Topic 1
\\
2. Topic 2"
}
\\
If you feel that there is no redundancy in the slide and it is concise, respond with the following sentence: 
\\

\textit{"Slides are concise and there is little to no
redundant information”.}
"
\end{boxA}

Similarly, as explained in section 6 \ref{sec:REFLEX}, models for other metrics are prompted appropriately. Notice that the prompting follows the exact same pattern as the pseudo explanations generated, to ensure consistency in baseline comparison.

\section{Inference Pipeline}
\begin{figure}[h]
    \centering
    \includegraphics[width=\linewidth]
    {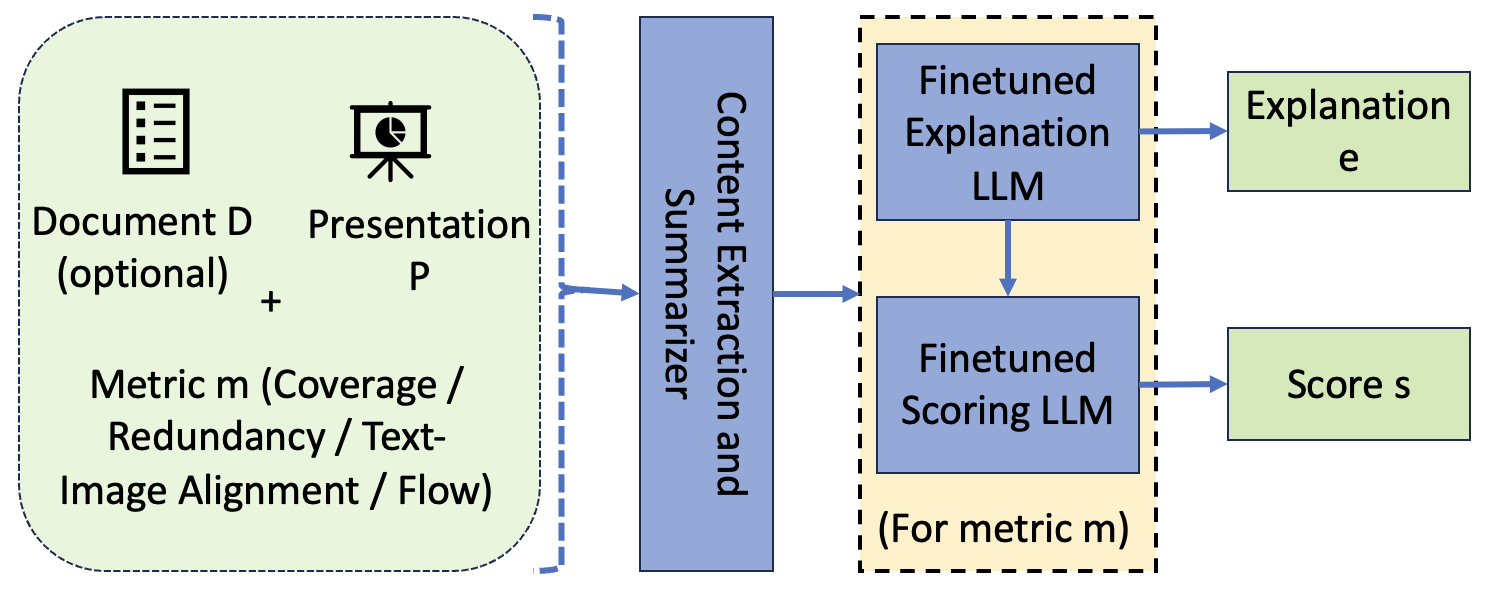}
    \caption{Inference pipeline of REFLEX}
    \label{fig:reflex_inference}
\end{figure}

Figure \ref{fig:reflex_inference} explains the inference pipeline of REFLEX.

\section{Human Annotation}

The human annotators selected for this study were a group of 10 machine learning researchers. All annotators were comfortable with the language and the topics present in the slides. The following instructions were given to them to score the slides (both positive and perturbed):

\begin{boxA}
You will evaluate a set of presentation slides based on four metrics: \textbf{Coverage}, \textbf{Redundancy}, \textbf{Text-Image Alignment}, and \textbf{Flow}. Both original and perturbed slides are provided for comparison. Use the 5-point Likert scale \textbf{[1 (Very Poor) - 5 (Excellent)]} for each metric.

\subsection*{1. Coverage}
Coverage assesses how well the slides capture key information from the source document. Consider:
\begin{itemize}
    \item Are key topics from the document included in the slides?
    \item Are important topics missing?
\end{itemize}
\textbf{Score:} High if most key topics are covered; low if important content is missing.

\subsection*{2. Redundancy}
Redundancy evaluates the presence of repetitive information. Consider:
\begin{itemize}
    \item Are slides free of unnecessary repetition?
    \item Could any slides be combined or removed?
\end{itemize}
\textbf{Score:} High if the presentation is concise; low if content is repeated unnecessarily.

\subsection*{3. Text-Image Alignment}
This metric measures the relevance of images to the text. Consider:
\begin{itemize}
    \item Do images complement and clarify the text?
    \item Are any images irrelevant to the slide's content?
\end{itemize}
\textbf{Score:} High if images and text are well-aligned; low if images are irrelevant.

\subsection*{4. Flow}
Flow assesses the logical progression of the slides. Consider:
\begin{itemize}
    \item Does the presentation have a smooth and logical flow?
    \item Are transitions between slides clear and cohesive?
\end{itemize}
\textbf{Score:} High if the flow is logical; low if the content feels disjointed.

\end{boxA}

And the following instructions were given to score the explanations generated by the model and the baselines.
\begin{boxA}
\subsection*{Scoring the Explanations}

In addition to scoring the slides, you will evaluate the \textbf{explanations} generated for each slide on a 5-point Likert scale \textbf{[1 (Very Poor) - 5 (Excellent)]}. Consider the following criteria:

\begin{itemize}
    \item \textbf{Clarity:} Is the explanation easy to understand?
    \item \textbf{Actionability:} Does the explanation provide clear steps or suggestions for improving the slide or presentation?
    \item \textbf{Relevance:} Is the explanation directly related to the issues identified in the slide?
\end{itemize}

\textbf{Score:} 
\begin{itemize}
    \item High if the explanation is clear, actionable, and relevant.
    \item Low if the explanation is vague, lacks actionable feedback, or is not related to the slide's issues.
\end{itemize}

\end{boxA}

\section{Inter-Annotator Agreement}\label{appen:inter-anno}
\begin{table}[h]
\centering
\resizebox{0.8\linewidth}{!}{
\renewcommand{\arraystretch}{0.8} 
\begin{tabular}{lcccc}
\toprule
\textbf{Metric} & \multicolumn{2}{c}{\textbf{$\rho$}} & \multicolumn{2}{c}{\textbf{$\tau$}} \\ 
\cmidrule(lr){2-3} \cmidrule(lr){4-5}
 & \textbf{mean$\uparrow$} & \textbf{max$\uparrow$} & \textbf{mean$\uparrow$} & \textbf{max$\uparrow$} \\ 
\midrule
\textbf{Coverage} & 0.57 & \textbf{0.61} & 0.56 & \textbf{0.58} \\
\textbf{Redundancy} & 0.54 & \textbf{0.63} & 0.57 & \textbf{0.62} \\
\textbf{Text-Img Ali.} & 0.63 & \textbf{0.67} & 0.61 & \textbf{0.66} \\
\textbf{Flow} & 0.43 & \textbf{0.51} & 0.39 & \textbf{0.47} \\
\bottomrule
\end{tabular}
}
\caption{Inter-Annotator Agreement: Mean and Maximum Correlations for Different Metrics}
\label{table:inter-anno}
\end{table}

\section{Hyperparameter Search}
\begin{table}[h]
\centering
\resizebox{0.4\textwidth}{!}{
\begin{tabular}{lcc}
\toprule
\textbf{Hyperparameter} & \textbf{Values} \\
\midrule
\textbf{Epochs} & \{1, 5, 10, 20, 30\} \\
\textbf{Batch Size} & \{16, 32, 64\} \\
\textbf{Learning Rates} & \{1e-3, 5e-4, 1e-4, 5e-3, 1e-5\} \\
\textbf{Dropout Rate} & \{0.1, 0.2, 0.3\} \\
\textbf{Weight Decay} & \{0, 0.01, 0.1\} \\
\bottomrule
\end{tabular}
}
\caption{Hyperparameters used for fine tuning.}
\label{table:hyperparameters}
\end{table}

\section{Sample Output from REFLEX}
\begin{figure*}[h]
    \centering
    \includegraphics[width=\linewidth]
    {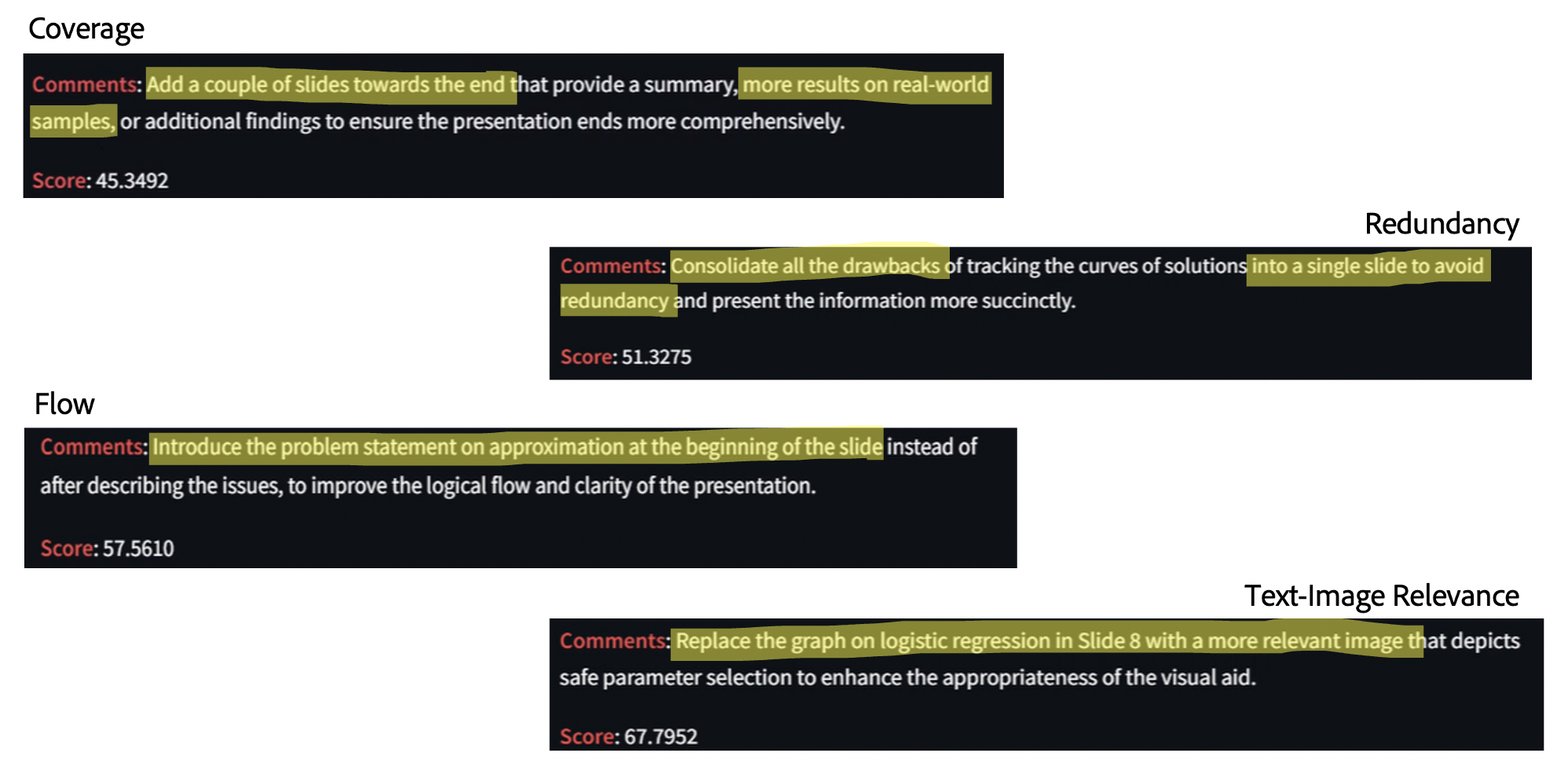}
    \caption{Sample output of actionable feedback from REFLEX}
    \label{fig:sample_output}
\end{figure*}

We have included a sample output of scores and actionable feedback from REFLEX in Figure \ref{fig:sample_output}.

\section{Related Works}\label{appen:related_work}
\label{sec:rel_works}

Many prior works use reference-based evaluation methods to assess target content, with model-free metrics such as BLEU \citep{papineni-etal-2002-bleu} and ROUGE \citep{lin-2004-rouge} being popular for measuring N-gram overlap between generated and reference texts. While these metrics effectively capture structural similarity, they fail to account for more nuanced aspects like semantic similarity. To address this, model-based metrics such as BERTScore \citep{bertscore} compute the maximum cosine similarity between BERT embeddings \citep{devlin2019bertpretrainingdeepbidirectional} of generated and reference texts. Multi-sentence texts can be evaluated using Sentence Mover’s Similarity (SMS) \citep{clark-etal-2019-sentence}, and MoverScore \citep{zhao2019moverscoretextgenerationevaluating} further enhances BERTScore with advanced aggregation strategies. Despite improvements, these methods are limited in capturing contextual nuances like sentence order and language style. Recent work has introduced LLM-based metrics, such as EVA-SCORE \citep{fan2024evascoreevaluationlongformsummarization}, which evaluates long-form summarization using atomic fact chains and document-level relation extraction, and \citep{luo2023chatgptfactualinconsistencyevaluator}, which uses ChatGPT to assess factual consistency. However, reference-based metrics still rely heavily on annotated data, limiting their broader applicability.


Reference-free metrics have gained significant popularity due to their ease of use. Approaches such as \citet{sun-nenkova-2019-feasibility,gao-etal-2020-supert,chen2018semanticqabasedapproachtext} employ regression models to align evaluations with human judgments. \citet{liu2023gevalnlgevaluationusing} introduces an LLM-based method that leverages chain-of-thought prompting to enhance alignment with human evaluations. Additionally, \citet{trainin2024covscoreevaluationmultidocumentabstractive, nguyen2024comparativestudyqualityevaluation} explore the usability of LLMs for evaluating tasks such as summarization and title-set generation.

\citet{villanovaaparisi2024readingorderindependentmetrics} introduces a set of reading order-independent metrics for evaluating information extraction in handwritten documents. \citet{scirè2024fenicefactualityevaluationsummarization} presents a factuality-oriented metric for text summarization, enhancing interpretability and efficiency through NLI-based alignment between source documents and extracted claims.

Recent works in language model evaluation focus on precision, efficiency, and robustness. \citet{li2024easyjudgeeasytousetoolcomprehensive} provides a lightweight, user-friendly model optimized with refined prompts for consistent evaluations. \citet{polo2024efficientmultipromptevaluationllms} estimates performance across multiple prompts, enabling robust metrics like top quantiles. \citet{lan2024criticeval} assesses critique abilities across diverse tasks using annotated references for reliability. These approaches highlight the need for scalable, multi-dimensional, and reliable evaluation methods for LLMs.

Most of these proposed methods perform well for evaluating relatively straightforward tasks, such as summarization and title-set generation, which involve only textual content. However, they fall short when assessing multi-modal content, such as presentations, where the relationship between images and text plays a crucial role. Additionally, these methods struggle to capture the nuances in visually complex documents that arise from intricate structural elements.

\end{document}